# Multi-modal Loop Closure Detection with Foundation Models in Severely Unstructured Environments

Laura Alejandra Encinar Gonzalez[1], John Folkesson[1], Rudolph Triebel[1,2] and Riccardo Giubilato[1]

*Abstract*— Robust loop closure detection is a critical component of Simultaneous Localization and Mapping (SLAM) algorithms in GNSS-denied environments, such as in the context of planetary exploration. In these settings, visual place recognition often fails due to aliasing and weak textures, while LiDAR-based methods suffer from sparsity and ambiguity. This paper presents MPRF, a multimodal pipeline that leverages transformer-based foundation models for both vision and LiDAR modalities to achieve robust loop closure in severely unstructured environments. Unlike prior work limited to retrieval, MPRF integrates a two-stage visual retrieval strategy with explicit 6-DoF pose estimation, combining DINOv2 features with SALAD aggregation for efficient candidate screening and SONATA-based LiDAR descriptors for geometric verification. Experiments on the S3LI dataset and S3LI Vulcano dataset show that MPRF outperforms state-of-the-art retrieval methods in precision while enhancing pose estimation robustness in low-texture regions. By providing interpretable correspondences suitable for SLAM back-ends, MPRF achieves a favorable trade-off between accuracy, efficiency, and reliability, demonstrating the potential of foundation models to unify place recognition and pose estimation. Code and models will be released at `github.com/DLR-RM/MPRF`.

## I. INTRODUCTION

Place recognition is critical to SLAM (Simultaneous Localization and Mapping) algorithms to function accurately in Global Navigation Satellite System (GNSS)-denied environments such as in the context of planetary exploration, where autonomous rovers must navigate unstructured, feature-less terrains. In such conditions, visual methods often fail due to repetitive textures or lack of distinctive landmarks, while geometry-only approaches are sensitive to sparsity and viewpoint variation [1].

Recent learning-based pipelines have improved place recognition by extracting more transferable descriptors and enabling large-scale retrieval. However, most existing approaches remain limited to retrieval-only, providing no explicit relative 6D pose estimates required for geometric loop closure. Similarly, while multimodal vision-LiDAR methods improve retrieval robustness and can sometimes provide coarse pose estimates, they predominantly output similarity scores and fall short of the metric 6-DoF transformations required for direct integration into SLAM back-ends [2]–[4].

In this paper, we present **MPRF** (Multimodal Place Recognition leveraging Foundation models), a unified pipeline for efficient loop closure detection in challenging unstructured environments. MPRF leverages transformer-based foundation models for both vision and LiDAR modalities, aggregates descriptors using scalable indexing, and refines candidates via patch-level similarity. Crucially, it extends beyond retrieval by estimating 6-DoF relative poses through geometric verification, bridging the gap between place recognition and SLAM.

*This work was supported by the Helmholtz Association project ARCHES (contract number ZT-0033) and project iFOODis (contract number KA2-HSC-06)
[1]Institute of Robotics and Mechatronics, German Aerospace Center (DLR), Weßling, Germany `fistname.lastname@dlr.de`
[2]Rudolph Triebel is also with Karlsruhe Institute of Technology (KIT), Faculty of..., Karlsruhe, Germany `rudolph.triebel@kit.de`

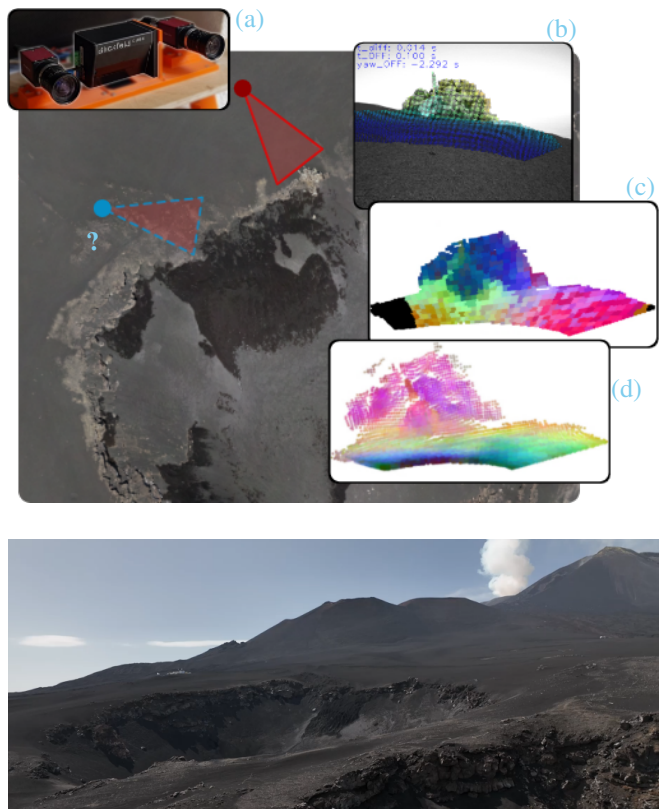

Fig. 1: **(top)**: Impressions and elements from the proposed multi-modal place recognition pipeline MPRF: (a) S3LI (Stereo Solid-State LiDAR Inertial) sensor setup used to gather multi-modal data, (b) Captured monochrome image of a large rock formation, and example projection of LiDAR pointcloud. (c) DinoV2 embeddings, PCA colored, projected on the LiDAR pointcloud, and (d) SONATA embeddings, PCA colored, computed from the pointcloud directly. **(bottom)**: Impression of the Cisternazza site, on the slopes of Mt. Etna, corresponding to the aerial imaging above.

The main contributions are:

- A multimodal pipeline that combines DINOv2-based

- image features and SONATA-based LiDAR descriptors, leveraging their large-scale pretraining to achieve strong generalization in planetary-like environments where domain-specific data is scarce.
- A two-stage retrieval strategy: efficient global screening with SALAD-aggregated descriptors, followed by refinement using multi-layer patch embeddings and cosine similarity.
- Integration of geometric verification into the retrieval process, yielding explicit 6-DoF relative poses via PnP+RANSAC and ICP.
- Experimental validation on planetary-analog datasets, showing that MPRF achieves robust loop closures in low-texture, unstructured terrains outperforming uni-modal and retrieval-only methods.

## II. RELATED WORK

### A. Visual Place Recognition

Early VPR methods relied on handcrafted descriptors such as SIFT [5], SURF [6], and ORB [7], which offered invariance to viewpoint and scale but struggled in low-texture or visually-ambiguous environments. The advent of deep learning enabled discriminative global representations, with NetVLAD [8] introducing a differentiable VLAD layer, DenseVLAD [9] addressing low-texture regions, and Patch-NetVLAD [10] incorporating geometric consistency. More recent pooling strategies such as GeM [11], region aggregation [12], and MLP-Mixer architectures like MixVPR [13] further improved compactness and retrieval accuracy.

Transformers have significantly advanced VPR by modeling global spatial relationships between distant image regions and enabling self-supervised pretraining. DINO and DINOv2 [14], [15] produce transferable patch-level descriptors without annotated data, while TransVPR [16] aggregates multi-level transformer features for both global and patch-level retrieval. Recent aggregation approaches such as DinoMix [17] and SALAD [18] combine transformer descriptors with optimal transport clustering, achieving robustness under strong viewpoint and illumination changes. Despite these advances, most VPR pipelines remain retrieval-focused and do not provide explicit 6-DoF constraints.

### B. LiDAR and Multi-Modal Place Recognition

LiDAR-based approaches exploit geometric structure for appearance-invariant localization. PointNet [19] and its derivatives, including PointNetVLAD [20] and LPD-Net [21], pioneered learned global 3D descriptors. Sparse convolutional methods such as MinkLoc3D [22] and its multi-modal extension MinkLoc++ [23] improved scalability, while attention-based models enhanced feature specificity [24]. Recently, SONATA [25] introduced a self-supervised point cloud transformer capable of multi-scale spatial reasoning, setting strong baselines for LiDAR-based retrieval.

To overcome modality-specific limitations, multimodal fusion methods combine complementary visual and LiDAR cues. Early work concatenated features [26], while later methods applied attention-based weighting, e.g., AdaFusion [27], or jointly modeled local and global multimodal features [3]. These strategies yield robust retrieval under environmental variability but often stop at similarity scoring without geometric verification.

### C. Pose Estimation

Classical approaches estimate relative pose from local correspondences using PnP with RANSAC [28], or via ICP [29] for point clouds. While accurate, they degrade under low texture or sparse geometry. Learning-based methods such as PoseNet [30] regress 6-DoF pose directly from images, but lack interpretability and underperform compared to geometry-based pipelines. Hybrid frameworks combining dense feature matchers (e.g., LoFTR [31]) with deep regressors [32] improve generalization, but the regression component provides no inlier statistics, limiting validation compared to correspondence-based pipelines. Scene coordinate regression approaches such as DSAC [33] integrate differentiable RANSAC into training, while recent foundation-model-based estimators (e.g., FoundPose [34]) leverage DINOv2 descriptors with PnP+RANSAC to achieve generalizable, training-free 6-DoF estimation. However, FoundPose targets object pose estimation and assumes access to accurate 3D models, which are not available in SLAM scenarios where the environment is unknown.

### D. 3D Point Descriptor Matching

For long-term LiDAR registration, handcrafted descriptors such as FPFH [35] and 3DMatch [36] are increasingly replaced by learning-based methods. Recent developments exploit lifting visual features from foundation models (e.g., DINOv2) onto 3D point clouds [37], enabling geometry-aware descriptors that outperform specialized LiDAR networks under domain shifts. This hybridization of vision and geometry represents a promising step toward robust place recognition in challenging scenes.

## III. METHODOLOGY

### A. System Overview

The proposed system integrates visual and LiDAR modalities to perform place recognition and relative pose estimation to establish loop closures for SLAM. The pipeline consists of two sequential stages: (i) **image retrieval** using transformer-based features aggregated via a learned VLAD variant, and (ii) **geometric verification and pose estimation** using multi-modal 3D descriptors. This design ensures both efficient candidate retrieval and explainable 6-DoF transformations suitable for SLAM integration. An overview of the proposed pipeline is shown in Fig. 2.

### B. Visual Feature Extraction and Aggregation

We employ **DINOv2** [15], a self-supervised Vision Transformer trained on large-scale unlabeled data, to extract patch-level features from input images. Two complementary screening stages are used to balance retrieval efficiency and spatial discrimination (Fig. 2).

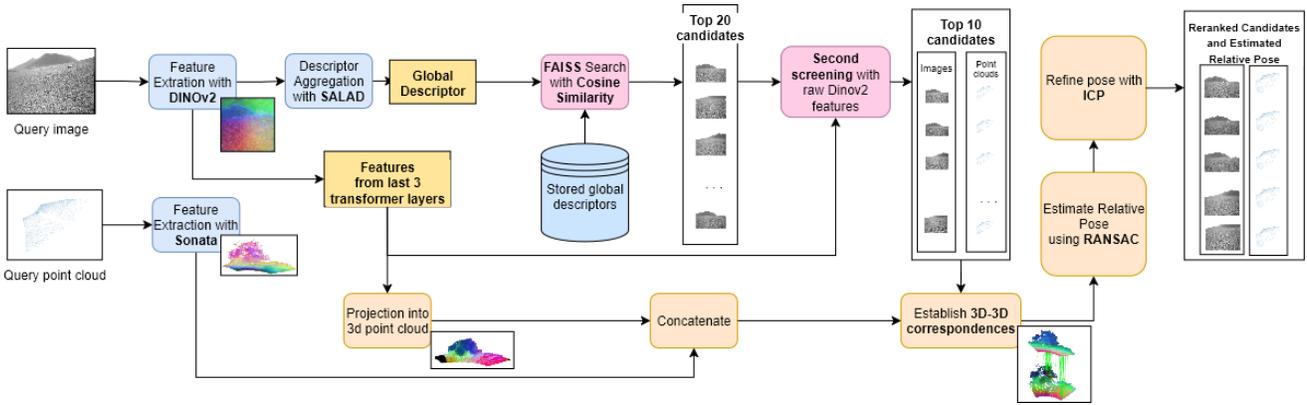

Fig. 2: Multi-modal localization pipeline combining global visual retrieval, DINOv2 feature matching, and LiDAR-based pose estimation.

**Global Retrieval:** Patch embeddings from the final transformer layer are aggregated into a compact global descriptor using SALAD (Sinkhorn Algorithm for Locally Aggregated Descriptors) [18]. SALAD performs VLAD-style clustering with learnable centers and a dustbin cluster to discard uninformative features. The resulting descriptors are compared using cosine similarity and indexed with FAISS (Facebook AI Similarity Search) to enable efficient approximate nearest-neighbor search [38], yielding a shortlist of candidate frames.

**Descriptor Refinement:** For the second stage, we concatenate embeddings from the last three layers and average them across patches, an approach designed to retain richer spatial cues while still producing a compact frame-level descriptor. These descriptors are compared between query and candidate images using cosine similarity, with FAISS providing efficient ranking [38]. This step emphasizes global spatial consistency while improving robustness to viewpoint and illumination changes. This two-stage strategy enables efficient large-scale retrieval while retaining discriminative power, ensuring reliable candidate selection in unstructured environments.

*C. Fine-Tuning of Visual Features*

To adapt the visual backbone to our specific test case of planetary-like scenes, we fine-tuned DINOv2 on the S3LI dataset [1], using the sequences *traverse_2*, *landmarks*, *crater*, *crater_inout*, and *mapping*. A total of 219,460 triplets were generated, split 80% for training and 20% for validation. Each triplet consists of an anchor, a positive (overlap $> 0.7$), and a negative (overlap $< 0.1$), with a temporal separation of at least 100 ms, where overlap is defined by the yaw-based angular alignment with a position correction (see IV-A). Data augmentation was applied on-the-fly to improve generalization, including random variations in brightness, contrast, saturation, and hue, as well as resized cropping. Training was conducted on a dual-GPU Quadro GV100 server using the AdamW optimizer (learning rate $1 \times 10^{-5}$, batch size 8), with a triplet margin loss ($m = 0.2$). Early stopping with a patience of two epochs was applied, and the best-performing checkpoint was retained. The training and validation loss curves (Fig. 3) indicate stable convergence without overfitting. The fine-tuned DINOv2 model was used in the final system both for patch-level embeddings and for the input to the SALAD aggregation module.

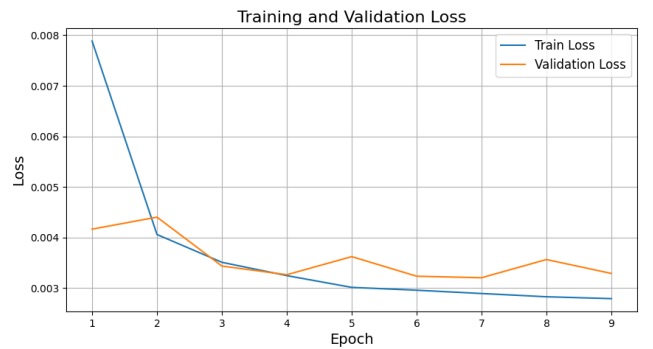

Fig. 3: Training and validation loss curves for DINOv2 fine-tuning, with early stopping after epoch 9.

*D. LiDAR Feature Extraction*

For geometric representation, given the scarcity of annotated planetary-like point cloud data, we leverage **SONATA** [25], a transformer-based backbone for 3D point cloud encoding trained on large-scale 3D datasets, exhibiting similar emergent properties as DINOv2. Each LiDAR frame is processed by SONATA into local descriptors that capture both geometric structure and semantic context, while remaining invariant to illumination and texture conditions. These descriptors are not used for global retrieval due to the increased computational cost and limited semantic value in sparse, repetitive terrains, but are crucial for pose estimation, where subtle geometric differences and depth information improve 6-DoF alignment once candidate matches are identified.

*E. Multi-Modal Feature Fusion*

Image patch embeddings are first projected into 3D using calibrated camera intrinsics and LiDAR depth measurements. Each 3D point is thus associated with both a visual descriptor and its corresponding LiDAR descriptor.

Before fusion, all descriptors are $\ell_2$-normalized. The paired features are then concatenated into a unified embedding. Candidate correspondences are computed using cosine similarity between fused descriptors, with a Hungarian assignment ensuring one-to-one matching. A similarity threshold of $0.90$ is applied to reject weak matches. This explicit matching strategy yields interpretable correspondences between image–LiDAR pairs.

*F. Pose Estimation*

The set of fused correspondences is used to estimate the relative pose between query and candidate frames (Fig. 2). We adopt RANSAC-based point-to-point registration from Open3D, with a tuned correspondence distance of $0.05$ meters. The algorithm iteratively samples minimal triplets of correspondences ($n = 3$) and optimizes a rigid transformation $\mathbf{T} \in SE(3)$ until convergence. Although the pipeline outputs full 6-DoF poses, evaluation focuses on yaw rotation and planar translation $(x, y)$, in line with the ground truth available in the S3LI dataset [1].

*G. Re-Ranking and Loop Closure Decision*

Candidate loop closures retrieved via FAISS are re-ranked according to the estimated relative pose distance between the query and candidate frames. This geometric consistency check ensures that matches correspond to spatially plausible loop closures. A loop closure is confirmed if RANSAC yields a valid 6-DoF transformation. No additional thresholds on inlier count or error metrics were applied, though alternative strategies (e.g., re-ranking by inlier count) could further improve robustness and are left for future work.

## IV. EXPERIMENTAL RESULTS

We evaluate MPRF on the S3LI dataset [1], a multimodal dataset that addresses scenarios which, for their visual and structural appearance, expose perceptual challenges such as the ones encountered in planetary-like settings. To avoid spatial leakage, the dataset was split into disjoint training and testing traverses, with evaluation performed on the *s3li_loops* and *traverse_1* sequences.

To further validate generalization capabilities, the place recognition component was also tested on validation sequences provided from the Vulcano extension of the same dataset, recorded on the homonymous island from the Aeolian islands, Sicily. The Vulcano sequences include more diverse environments than the original Mt. Etna ones, such as basaltic and iron-rich rock formations, old lava channels, dry vegetation, and waterfront areas. In our evaluation, we specifically used the *moon_lake*, *waterfront*, and *lava_tunnel* sequences, see Fig. 4 for a collection of visual impressions.

*A. Data Setup*

Pre-processing of the data was performed using the S3LI toolkit [39], which synchronizes input from the visual and LiDAR modalities, computes ground-truth overlap between data samples, and provides visualization for validation. Ground-truth overlap between image–scan pairs relies on D-GNSS measurements, and is computed as the product of two terms: (i) the angular alignment, measured from the yaw difference within the horizontal field of view of the cameras, and (ii) a position-based correction that down-weights pairs with large lateral or forward displacements. With reference to the definitions from [39], an overlap threshold of $\tau_o = 0.6$ defines true matches: retrieval candidates are considered positive if their overlap score exceeds $\tau_o$, and negative otherwise. With respect to the pose estimation task, the evaluation focuses on yaw and translation errors.

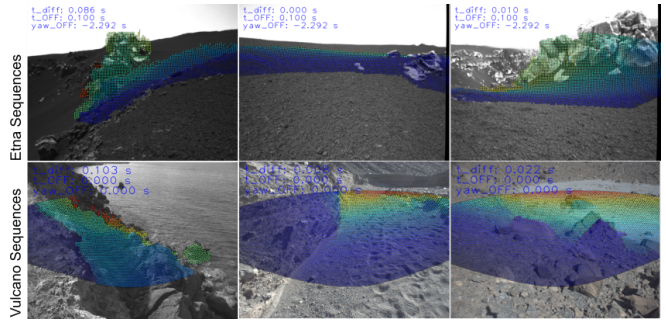

Fig. 4: Visual impressions from the S3LI dataset, with LiDAR pointcloud overlays. Top: Selection from Etna sequences. Bottom: Selection from the Vulcano validation sequences

*B. Implementation Details*

Input images are resized to 224 × 224 and processed with a ViT-B/14 DINOv2 backbone, producing 768-dim patch embeddings. For the global retrieval phase, features are aggregated with SALAD into 8192-dim descriptors. For the descriptor refinement stage, patch descriptors are obtained by concatenating the last three transformer layers (2304-dim). For pose estimation, LiDAR scans are encoded with SONATA into 512-dim voxel descriptors, providing a structured set of local geometric features.

Runtime is measured end-to-end, including feature extraction, descriptor aggregation, nearest-neighbor search, and re-ranking for retrieval; and feature projection, correspondence search, and RANSAC optimization for pose estimation. Reported times correspond to the average per query. All experiments are run on an NVIDIA RTX A4000 GPU.

*C. Retrieval Performance*

Table I reports retrieval precision at top-$k$ and average runtime across visual, LiDAR, and multimodal baselines. The proposed pipeline (MPRF) achieves the highest overall performance, with the best configuration (fine-tuned DINOv2 + pretrained SALAD, MPRF-PF) reaching 75.7% Precision@1 while maintaining retrieval time under 500 ms, measured as the end-to-end per-query duration (see IV-B).

A key observation is that LiDAR-only methods contribute little to retrieval in this setting. PointNetVLAD, despite its high computational cost (∼4 s per query), performs poorly with only 7.9% Precision@1. Similarly, the multimodal baseline MinkLoc++ underperforms compared to visual-only approaches such as NetVLAD, TransVPR, or DINOv2-based descriptors, showing that geometric information in sparse planetary-like point clouds does not improve large-scale retrieval. This analysis motivated our decision to exclude LiDAR from the retrieval stage of MPRF, reserving it instead for pose estimation where geometric cues are more impactful.

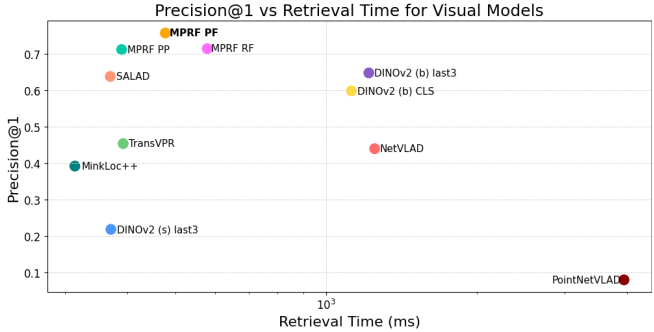

Fig. 5: Trade-off between retrieval time (in milliseconds) and Precision@1 for all evaluated methods. Top-left positions represent models that are both accurate and efficient. The proposed method (SALAD+DINO- ft) achieves the best balance, outperforming traditional baselines.

Within visual methods, two trends are evident. First, DINOv2 descriptors outperform classical CNN-based aggregation (NetVLAD) and transformer-based aggregation (TransVPR), especially when using multi-layer patch embeddings (64.7% Precision@1). Second, the SALAD aggregation module proves highly effective: when combined with fine-tuned DINOv2, it yields the strongest overall balance between accuracy and efficiency. Notably, attempts to retrain SALAD on the target domain degraded performance, suggesting that the pretrained clustering space is more stable and generalizable.

On the Vulcano validation sequences, see Table II, MPRF - PF maintains high performance (78.3% Precision@1) beating the compared top-performing methods on the previous Etna sequences, demonstrating robust generalization to unseen volcanic terrains. Interestingly, both DINOv2 and SALAD perform slightly better on these sequences, likely due to more informative images, containing vegetation, varied rock formations, and water, enhancing discriminative feature extraction and aggregation.

In summary, retrieval in planetary-like environments benefits most from robust self-supervised visual descriptors with adaptive aggregation, while LiDAR data provides little added value at this stage. Fig. 5 illustrates the trade-off between accuracy and runtime, where MPRF achieves the most favorable balance among all evaluated methods.

### D. Pose Estimation Performance

Table III summarizes yaw and translation errors, number of valid poses, and runtime for different families of methods. Handcrafted descriptors such as FPFH combined with RANSAC yield a pose estimate for every query, since no correspondences are filtered out. However, the resulting transformations are often inaccurate, with yaw errors exceeding 46° and translation errors above 8 m, while runtime surpasses 12 seconds per query. This makes FPFH unsuitable for online SLAM despite its apparent completeness in producing estimates.

Dense matching approaches trade coverage for accuracy: LoFTR achieves low translation errors (4.13 m in $x$, 6.92 m in $y$), yet yields valid poses for fewer than half the pairs, reflecting the difficulty of purely visual correspondence search in feature-sparse environments. Like other visual-only methods, it produces a 5D transformation (rotation and translation without scale) unless external depth information is used.

In contrast, patch-level DINO features and SONATA-only descriptors provide a transformation for most queries with lower errors than handcrafted FPFH, but translation accuracy remains limited. A projection-based fusion baseline (DINO-LiDAR), where visual descriptors are simply lifted into 3D using LiDAR depth, does not yield improvements: the features remain appearance-driven and fail to exploit the structural cues in the point cloud. In contrast, our proposed fusion combines SONATA's geometric descriptors with DINO embeddings, producing multimodal correspondences that integrate complementary appearance and structural information. This richer representation leads to more reliable pose estimation in feature-sparse environments.

Regression-based estimators stand out in efficiency. Reloc3r achieves the lowest yaw error (8.15°) and runs in just 134 ms per query, but as a visual-only method, it also outputs 5D poses and provides no explicit correspondences, limiting interpretability and downstream validation.

Our proposed descriptor-level multimodal fusion strikes a balance between accuracy, robustness, and interpretability. By integrating DINOv2 and SONATA descriptors, it achieves competitive angular accuracy (8.20° yaw error) while delivering valid poses for all candidate pairs. Although runtime (3.1 s per query) is higher than visual-only approaches, it remains well below handcrafted baselines and can be optimized further. Most importantly, unlike regression, this pipeline provides explicit geometric verification, ensuring that loop closures are both accurate and explainable for reliable integration into SLAM.

### E. Pose Estimation Reliability: Threshold-Based Analysis.

We analyze the proportion of poses with yaw and translation errors within predefined thresholds (Tables IV, V and VI) to assess prediction consistency in feature-sparse conditions.

From Table IV, over 69% of poses estimated by the proposed model fall within 10° of the ground truth, closely matching Reloc3r's performance. While performance at tighter thresholds (2° and 3°) is slightly lower, the results indicate that the model provides consistent and reliable angular estimates.

Similarly, Tables V and VI show that 42.44% and 36.99% of poses have translation errors below 5 m in X and Y, re-spectively, closely matching Reloc3r, while LoFTR performs better at tighter thresholds (1–3 m).

TABLE I: Image Retrieval Results on the S3LI dataset: Precision at Top-$k$ and Average Retrieval Time

| Model | Modality | Precision@1 | Precision@5 | Precision@10 | Time (ms) |
|---|---|---|---|---|---|
| PointNetVLAD | L | 7.93 | 7.86 | 7.30 | 3942.52 |
| MinkLoc++ | V-L | 39.17 | 34.61 | 31.90 | 314.19 |
| NetVLAD | V | 43.95 | 42.37 | 40.33 | 1249.89 |
| TransVPR | V | 45.34 | 43.25 | 0.4126 | 392.39 |
| DINOv2 (s) (last 3 layers) | V | 21.79 | 20.33 | 18.97 | 370.87 |
| DINOv2 (b) (CLS Token) | V | 59.82 | 54.71 | 51.61 | 1123.11 |
| DINOv2 (b) (last 3 layers) | V | 64.74 | 60.81 | 58.38 | 1216.88 |
| SALAD (pretrained) | V | 63.78 | 70.48 | 67.92 | 369.82 |
| MPRF-PP (pretrained SALAD + pretrained DINOv2) | V | 71.16 | 68.97 | 67.95 | 389.71 |
| MPRF-RF (retrained SALAD + fine-tuned DINOv2) | V | 71.41 | 69.65 | 67.12 | 578.21 |
| **MPRF-PF (pretrained SALAD + fine-tuned DINOv2)** | V | **75.69** | **73.32** | **70.90** | 476.57 |

TABLE II: Image Retrieval Results on the S3LI Vulcano dataset: Precision at Top-$k$

| Model | P@1 | P@5 | P@10 |
|---|---|---|---|
| DINOv2 (b) (last 3 layers) | 0.6797 | 0.6510 | 0.6235 |
| SALAD (pretrained) | 0.7786 | 0.7546 | 0.7300 |
| **MPRF-PF** | **0.7834** | **0.7632** | **0.7497** |

This threshold-based analysis confirms that the multimodal fusion strategy yields not only accurate average pose estimates but also robust and reliable predictions suitable for downstream SLAM tasks, where even occasional large deviations can severely impact trajectory consistency.

*F. Ablation Studies*

We performed ablation experiments to assess the effect of backbone tuning, feature aggregation, and multimodal fusion. These experiments include comparisons between (i) pretrained DINOv2 + pretrained SALAD, (ii) fine-tuned DINOv2 + pretrained SALAD, and (iii) variants with retrained SALAD. This analysis provides insight into the design choices underlying MPRF.

Fine-tuning DINOv2 on the Etna dataset improved Precision@1 from 71.2% to 75.7%, highlighting the value of domain adaptation in planetary-like terrain where generic self-supervised features are insufficiently specific.

Different variants of the visual backbone were also evaluated within the retrieval pipeline. Using only the CLS token reduced performance (59.8% Precision@1), while aggregating patch descriptors from the last three layers improved accuracy to 64.7%. PCA visualizations (Fig. 6) illustrate that the three layers encode complementary spatial cues, justifying their aggregation for candidate refinement.

Retraining SALAD on the target dataset degraded performance (71.4% vs. 75.7% Precision@1), indicating the pretrained clustering space is more stable with limited data. Finally, we evaluated SONATA descriptors and DINO patch features independently. Both unimodal variants frequently resulted in pose errors above acceptable thresholds, limiting their suitability for loop closure. In contrast, their fusion significantly reduced yaw error and improved robustness in low-texture regions (Fig. 7). This confirms that geometry provides complementary cues for pose estimation when combined with visual embeddings.

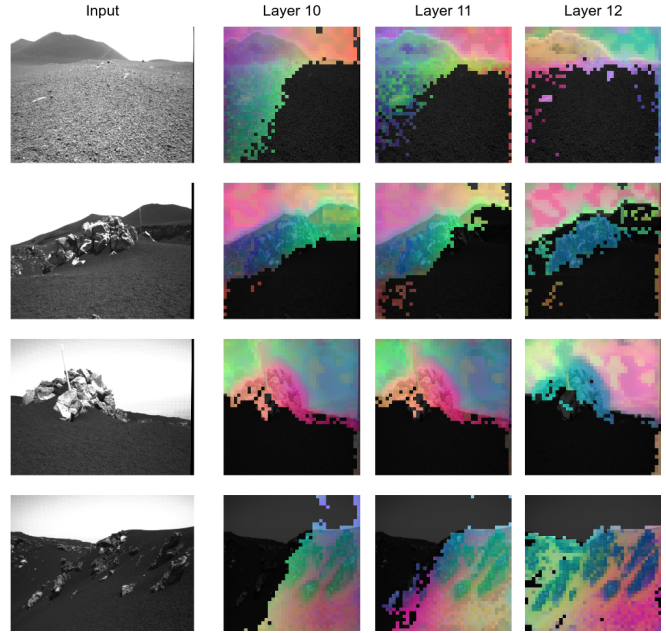

Fig. 6: Examples from the S3LI dataset (first column) and PCA visualizations of patch embeddings from the last three transformer layers of DINOv2 (columns 2–4). To improve interpretability, less informative regions were removed by discarding patches with a negative score along the first PCA component. The visualizations highlight how successive layers encode different spatial cues, motivating the use of multi-layer aggregation for robust candidate refinement.

V. CONCLUSIONS

We introduced MPRF, a multimodal pipeline combining **visual and LiDAR foundation models** for loop closure in unstructured environments. Unlike retrieval-only approaches, it integrates 6-DoF pose estimation, bridging place recognition and geometric loop closure, while leveraging large-scale priors to **reduce task-specific training** needs.

Experiments on S3LI highlighted three findings. First, visual foundation models excelled at retrieval: fine-tuning

TABLE III: Pose Estimation Results: Average Errors, Total Poses Estimated, and Inference Time. Visual-only methods marked with * (LoFTR, DINO-Patches, Reloc3r) return 5-DoF poses (rotation + translation up to scale)

| Model | Modality | Yaw Error (°) | DX Error (m) | DY Error (m) | Poses Estimated | Time (ms) |
|---|---|---|---|---|---|---|
| FPFH + RANSAC | L | 46.82 | 8.23 | 14.27 | **1560** | 12233.82 |
| DINO-LiDAR + RANSAC | V-L | 25.10 | 8.40 | 14.27 | **1560** | 5686.82 |
| LoFTR + RANSAC* | V | 11.40 | **4.13** | **6.92** | 744 | 249.36 |
| DINO-Patches + RANSAC* | V | 17.13 | 8.06 | 14.05 | 1353 | 948.30 |
| SONATA + RANSAC | L | 16.36 | 8.86 | 15.30 | 1066 | 3572.05 |
| Reloc3r* | V | **8.15** | 8.31 | 14.19 | **1560** | **133.76** |
| **MPRF (DINO + SONATA)** | V-L | 8.20 | 8.44 | 14.24 | **1560** | 3114.33 |

TABLE IV: Percentage of Estimated Poses with Yaw Error Below Thresholds.

| Model | < 2° | < 3° | < 5° | < 10° |
|---|---|---|---|---|
| LoFTR + RANSAC | 13.31 | 20.03 | 32.66 | 60.48 |
| Reloc3r (2025) | 16.22 | 25.00 | 41.60 | 71.22 |
| **MPRF** | 14.29 | 22.56 | 39.10 | 69.94 |

TABLE V: Percentage of Estimated Poses with Translation Error in X Below Thresholds.

| Model | < 1m | < 2m | < 3m | < 5m | < 10m |
|---|---|---|---|---|---|
| LoFTR + RANSAC | 22.58 | 36.83 | 51.48 | 72.85 | 91.67 |
| Reloc3r (2025) | 9.36 | 18.14 | 26.47 | 44.49 | 66.92 |
| **MPRF** | 8.65 | 18.21 | 27.37 | 42.44 | 65.45 |

TABLE VI: Percentage of Estimated Poses with Translation Error in Y Below Thresholds.

| Model | < 1m | < 2m | < 3m | < 5m | < 10m |
|---|---|---|---|---|---|
| LoFTR + RANSAC | 19.35 | 34.68 | 43.28 | 56.99 | 79.03 |
| Reloc3r (2025) | 11.09 | 20.60 | 28.33 | 37.63 | 57.05 |
| **MPRF** | 11.09 | 20.38 | 27.88 | 36.99 | 57.12 |

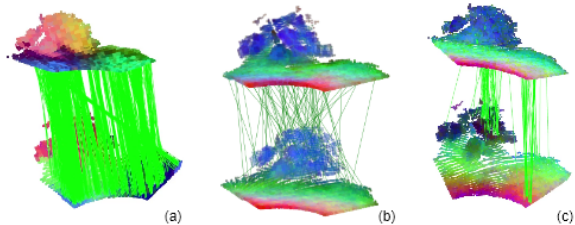

Fig. 7: Qualitative comparison of point correspondences. (a) DINOv2 visual features alone produce many incorrect matches due to the absence of geometric cues. (b) SONATA 3D features alone fail in low-texture or structurally repetitive areas. (c) The proposed fusion of DINOv2 and SONATA features yields more accurate, consistent, and less noisy correspondences, illustrating the complementary strengths of appearance and geometry.

DINOv2 on planetary-like data improved accuracy, and its combination with SALAD provided the best balance of efficiency and discriminative power, while LiDAR-only descriptors provided little benefit at this stage. Second, LiDAR geometry was critical for pose estimation: unimodal DINOv2 or SONATA descriptors often produced large errors, but their **fusion improved robustness in low-texture regions**, demonstrating the complementarity of appearance and structure. Third, interpretability and reliability remain essential: unlike regression-based methods, MPRF offers **explicit geometric verification** through correspondences that can be validated in SLAM back-ends, with threshold-based analysis confirming compact error distributions and fewer catastrophic outliers.

In summary, MPRF advances the integration of visual and LiDAR foundation models into a unified loop closure pipeline, achieving a balance of retrieval efficiency, pose estimation accuracy, and explainability. Future work will target faster pose estimation and integration into multimodal SLAM systems.

REFERENCES

[1] R. Giubilato, W. Sturzl, A. Wedler, and R. Triebel, "Challenges of SLAM in Extremely Unstructured Environments: The DLR Planetary Stereo, Solid-State LiDAR, Inertial Dataset," *IEEE Robotics and Automation Letters*, vol. 7, no. 4, pp. 8721–8728, 10 2022.

[2] C. Chen, B. Wang, C. X. Lu, N. Trigoni, and A. Markham, "A Survey on Deep Learning for Localization and Mapping: Towards the Age of Spatial Machine Intelligence." [Online]. Available: https://github.com/changhao-chen/deep-

[3] A. Garcia-Hernandez, R. Giubilato, K. H. Strobl, J. Civera, and R. Triebel, "Unifying Local and Global Multimodal Features for Place Recognition in Aliased and Low-Texture Environments," *Proceedings - IEEE International Conference on Robotics and Automation*, pp. 3991–3998, 3 2024. [Online]. Available: https://arxiv.org/abs/2403.13395v1

[4] J. Komorowski, M. W. Wysoczańska, and T. Trzcinski, "MinkLoc++: Lidar and Monocular Image Fusion for Place Recognition." [Online]. Available: https://github.com/jac99/MinkLocMultimodal

[5] D. G. Lowe, "Distinctive image features from scale-invariant keypoints," *International Journal of Computer Vision*, vol. 60, no. 2, pp. 91–110, 11 2004.

[6] H. Bay, T. Tuytelaars, and L. Van Gool, "SURF: Speeded up robust features," *Lecture Notes in Computer Science (including subseries Lecture Notes in Artificial Intelligence and Lecture Notes in Bioinformatics)*, vol. 3951 LNCS, pp. 404–417, 2006.

[7] E. Rublee, V. Rabaud, K. Konolige, and G. Bradski, "ORB: An efficient alternative to SIFT or SURF," *Proceedings of the IEEE International Conference on Computer Vision*, pp. 2564–2571, 2011.

[8] R. Arandjelovic, P. Gronat, A. Torii, T. Pajdla, and J. Sivic, "NetVLAD: CNN architecture for weakly supervised place recognition," *IEEE Transactions on Pattern Analysis and Machine Intelligence*, vol. 40, no. 6, pp. 1437–1451, 11 2015. [Online]. Available: https://arxiv.org/abs/1511.07247v3

[9] F. Magliani, T. Fontanini, and A. Prati, "A Dense-Depth Representation for VLAD descriptors in Content-Based Image Retrieval," Tech. Rep., 8 2018. [Online]. Available: http://implab.ce.unipr.it

[10] S. Hausler, S. Garg, M. Xu, M. Milford, and T. Fischer, "Patch-NetVLAD: Multi-Scale Fusion of Locally-Global Descriptors for Place Recognition," *IEEE Conference on Computer Vision and Pattern Recognition (CVPR) (2021)*, 2021. [Online]. Available:

http://ieeexplore.ieee.org

[11] "Generalized Mean Pooling Explained — Papers With Code." [Online]. Available: https://paperswithcode.com/method/generalized-mean-pooling

[12] M. Teichmann, A. Araujo, and M. A. Zhu Jack Sim Google, "Detect-to-Retrieve: Efficient Regional Aggregation for Image Search."

[13] A. Ali-Bey, B. Chaib-Draa, and P. Giguere, "MixVPR: Feature Mixing for Visual Place Recognition," *Proceedings - 2023 IEEE Winter Conference on Applications of Computer Vision, WACV 2023*, pp. 2997–3006, 3 2023. [Online]. Available: https://arxiv.org/abs/2303.02190v1

[14] M. Caron, H. Touvron, I. Misra, H. Jegou, J. Mairal, P. Bojanowski, and A. Joulin, "Emerging Properties in Self-Supervised Vision Transformers," *Proceedings of the IEEE International Conference on Computer Vision*, pp. 9630–9640, 4 2021. [Online]. Available: https://arxiv.org/abs/2104.14294v2

[15] M. Oquab, T. Darcet, T. Moutakanni, H. V. Vo, M. Szafraniec, V. Khalidov, P. Fernandez, D. Haziza, F. Massa, A. El-Nouby, M. Assran, N. Ballas, W. Galuba, R. Howes, P.-Y. Huang, S.-W. Li, I. Misra, M. Rabbat, V. Sharma, G. Synnaeve, H. Xu, H. Jegou, J. Mairal, P. Labatut, A. Joulin, and P. Bojanowski, "DINOv2: Learning Robust Visual Features without Supervision," 4 2023. [Online]. Available: https://arxiv.org/abs/2304.07193v2

[16] R. Wang, Y. Shen, W. Zuo, S. Zhou, and N. Zheng, "TransVPR: Transformer-based place recognition with multi-level attention aggregation," *Proceedings of the IEEE Computer Society Conference on Computer Vision and Pattern Recognition*, vol. 2022-June, pp. 13 638–13 647, 1 2022. [Online]. Available: https://arxiv.org/abs/2201.02001v4

[17] G. Huang, Y. Zhou, X. Hu, C. Zhang, L. Zhao, W. Gan, and M. Hou, "DINO-Mix: Enhancing Visual Place Recognition with Foundational Vision Model and Feature Mixing," 11 2023. [Online]. Available: https://arxiv.org/abs/2311.00230v2

[18] S. Izquierdo and J. Civera, "Optimal Transport Aggregation for Visual Place Recognition," 11 2023. [Online]. Available: https://arxiv.org/abs/2311.15937v2

[19] C. R. Qi, H. Su, K. Mo, and L. J. Guibas, "PointNet: Deep Learning on Point Sets for 3D Classification and Segmentation," *Proceedings - 30th IEEE Conference on Computer Vision and Pattern Recognition, CVPR 2017*, vol. 2017-January, pp. 77–85, 12 2016. [Online]. Available: https://arxiv.org/abs/1612.00593v2

[20] M. A. Uy and G. H. Lee, "PointNetVLAD: Deep Point Cloud Based Retrieval for Large-Scale Place Recognition," *Proceedings of the IEEE Computer Society Conference on Computer Vision and Pattern Recognition*, pp. 4470–4479, 4 2018. [Online]. Available: https://arxiv.org/abs/1804.03492v3

[21] Z. Liu, S. Zhou, C. Suo, Y. Liu, P. Yin, H. Wang, and Y.-H. Liu, "LPD-Net: 3D Point Cloud Learning for Large-Scale Place Recognition and Environment Analysis."

[22] J. Komorowski, "MinkLoc3D: Point Cloud Based Large-Scale Place Recognition," *Proceedings - 2021 IEEE Winter Conference on Applications of Computer Vision, WACV 2021*, pp. 1789–1798, 11 2020. [Online]. Available: https://arxiv.org/abs/2011.04530v1

[23] J. Komorowski, M. Wysoczanska, and T. Trzcinski, "MinkLoc++: Lidar and Monocular Image Fusion for Place Recognition," *Proceedings of the International Joint Conference on Neural Networks*, vol. 2021-July, 4 2021. [Online]. Available: https://arxiv.org/abs/2104.05327v2

[24] M.-H. Guo, J.-X. Cai, Z.-N. Liu, T.-J. Mu, R. R. Martin, S.-M. Hu, and T. Author, "PCT: Point cloud transformer," vol. 7, no. 2, pp. 187–199, 2021. [Online]. Available: https://doi.org/10.1007/s41095-021-0229-5

[25] X. Wu, D. Detone, D. Frost, T. Shen, C. Xie, N. Yang, J. Engel, R. Newcombe, H. Zhao, and J. Straub, "Sonata: Self-Supervised Learning of Reliable Point Representations Semantic Awareness Perception Self-distillation PCA K-means Dense Matching Sparse Matching," Tech. Rep., 2025. [Online]. Available: https://github.com/facebookresearch/sonata

[26] G. Wang, Y. Zheng, Y. Wu, Y. Guo, Z. Liu, Y. Zhu, W. Burgard, and H. Wang, "End-to-end 2D-3D Registration between Image and LiDAR Point Cloud for Vehicle Localization," 7 2025. [Online]. Available: http://arxiv.org/abs/2306.11346

[27] H. Lai, P. Yin, and S. Scherer, "AdaFusion: Visual-LiDAR Fusion with Adaptive Weights for Place Recognition," *IEEE Robotics and Automation Letters*, vol. 7, no. 4, pp. 12 038–12 045, 11 2021. [Online]. Available: https://arxiv.org/abs/2111.11739v1

[28] M. A. Fischler and R. C. Bolles, "Random sample consensus," *Communications of the ACM*, vol. 24, no. 6, pp. 381–395, 6 1981. [Online]. Available: https://dl.acm.org/doi/10.1145/358669.358692

[29] "ICP registration - Open3D 0.19.0 documentation." [Online]. Available: https://www.open3d.org/docs/release/tutorial/pipelines/icp_registration.html

[30] A. Kendall, M. Grimes, and R. Cipolla, "PoseNet: A Convolutional Network for Real-Time 6-DOF Camera Relocalization," 5 2015. [Online]. Available: http://arxiv.org/abs/1505.07427

[31] J. Sun, Z. Shen, Y. Wang, H. Bao, and X. Zhou, "LoFTR: Detector-Free Local Feature Matching with Transformers," Tech. Rep., 4 2021. [Online]. Available: https://zju3dv.github.io/loftr/.

[32] S. Dong, S. Wang, S. Liu, L. Cai, Q. Fan, J. Kannala, and Y. Yang, "Reloc3r: Large-Scale Training of Relative Camera Pose Regression for Generalizable, Fast, and Accurate Visual Localization," 12 2024. [Online]. Available: https://arxiv.org/abs/2412.08376v2

[33] E. Brachmann, A. Krull, S. Nowozin, J. Shotton, F. Michel, S. Gumhold, and C. Rother, "DSAC - Differentiable RANSAC for Camera Localization," *Proceedings - 30th IEEE Conference on Computer Vision and Pattern Recognition, CVPR 2017*, vol. 2017-January, pp. 2492–2500, 11 2016. [Online]. Available: https://arxiv.org/abs/1611.05705v4

[34] E. P. Örnek, Y. Labbé, B. Tekin, L. Ma, C. Keskin, C. Forster, and T. Hodan, "FoundPose: Unseen Object Pose Estimation with Foundation Features," 11 2023. [Online]. Available: https://arxiv.org/abs/2311.18809v2

[35] R. B. Rusu, N. Blodow, and M. Beetz, "Fast Point Feature Histograms (FPFH) for 3D Registration," *Proceedings - IEEE International Conference on Robotics and Automation*, pp. 3212–3217, 2009.

[36] A. Zeng, S. Song, M. Nießner, M. Fisher, J. Xiao, and T. Funkhouser, "3DMatch: Learning Local Geometric Descriptors from RGB-D Reconstructions," Tech. Rep. [Online]. Available: http://3dmatch.cs.princeton.eduhttp://3dmatch.cs.princeton.edu.

[37] N. Vödisch, G. Cioffi, M. Cannici, W. Burgard, and D. Scaramuzza, "LiDAR Registration with Visual Foundation Models," Tech. Rep. [Online]. Available: https://vfm-registration.cs.uni-freiburg.de.

[38] "Faiss." [Online]. Available: https://ai.meta.com/tools/faiss/

[39] "GitHub - DLR-RM/s3li-toolkit." [Online]. Available: https://github.com/DLR-RM/s3li-toolkit